\documentclass[twoside]{article}

\usepackage{mathrsfs}
\usepackage[numbers]{natbib}
\usepackage{caption}
\usepackage{booktabs}

\usepackage{floatrow}
\newfloatcommand{capbtabbox}{table}[][\FBwidth]

\usepackage[utf8]{inputenc} 
\usepackage[T1]{fontenc}    
\usepackage{hyperref}       
\usepackage{url}            
\usepackage{booktabs}       
\usepackage{amsfonts}       
\usepackage{nicefrac}       
\usepackage{microtype}      

\usepackage{amsmath, latexsym}
\usepackage{amscd}
\usepackage{amsbsy}
\usepackage{amssymb}
\usepackage{amsthm}
\usepackage{todonotes}
\DeclareMathOperator{\Risk}{Risk}
\DeclareMathOperator{\Bias}{Bias}
\DeclareMathOperator{\Var}{Var}

\newtheorem{theorem}{Theorem}
\newtheorem{corollary}[theorem]{Corollary}

{
	\theoremstyle{plain}
	\newtheorem{assumption}{Assumption}
}

\DeclareMathOperator{\Unif}{Unif}

\newcommand{\E}{{\mathbb E}}
\DeclareMathOperator{\argmin}{argmin}

\newcommand{\norm}[1]{\left\lVert #1 \right\rVert}

\usepackage[accepted]{aistats2018}
\begin{document}

%

%

\twocolumn[

\aistatstitle{Policy Evaluation and Optimization with Continuous Treatments}

 \aistatsauthor{ Nathan Kallus \And Angela Zhou }

\aistatsaddress{ Cornell University and Cornell Tech  \And  Cornell University } 
]

\vspace{-7pt}
\begin{abstract}
We study the problem of policy evaluation and learning from batched contextual bandit data when treatments are
continuous, going beyond previous work on discrete treatments.
Previous work for discrete treatment/action spaces focuses on inverse probability weighting (IPW)
and doubly robust (DR) methods that use a rejection sampling approach for evaluation
and the equivalent weighted classification problem for learning.
In the continuous setting, this reduction fails as we would almost surely
reject all observations.
To tackle the case of continuous treatments, we extend the
IPW and DR approaches to the continuous setting using a kernel function
that leverages treatment proximity to attenuate discrete rejection.
Our policy estimator is consistent and we
characterize the optimal bandwidth.
The resulting \textit{continuous policy optimizer} (CPO) approach using our estimator achieves convergent regret and
approaches the best-in-class policy for learnable policy classes. We demonstrate that the estimator performs well and, in particular,
outperforms a discretization-based benchmark.
We further study the performance of our policy optimizer in a
case study on personalized dosing based on a dataset of Warfarin
patients, their covariates, and final therapeutic doses.
Our learned policy outperforms benchmarks and nears
the oracle-best linear policy.
\end{abstract}

\section{Introduction}

Personalization is a key feature of modern decision making in a variety of contexts
and learning how to personalize is a central problem in machine learning.
In targeted advertising, learning systems observe data about incoming users such as their
market segment, decide which ad to display to the individual user, and observe whether or not
the user clicks on the ad. In personalized medicine, medical history, demographics, and
genetics of patients may be leveraged to administer individually tailored treatments.
Contextual-bandit algorithms, where the learning system repeatedly takes an action for a context
(feature vector) and observes the corresponding outcome, and other
randomized experiments such as A/B tests, are the gold standards for comparing the
efficacy of different policies and learning the best one.
However, in practice, it can be prohibitively costly, risky, or impossible to
collect new data through experimentation.
Off-policy evaluation and optimization is the problem of assessing and optimizing
new personalized decision policies based on observational data collected under
other historical policies.

The existing literature on policy evaluation has primarily considered only discrete
action spaces, where the learning system chooses one of $K$ treatments for each unit \citep{bl09,dell2014,lwlw11}, except for model-based approaches for policy evaluation \citep{kgdh2015}. In many
important applications, however, the treatment is a continuous variable. For example, the
dosage of a medical drug is continuous and, by using personalized dosing policies, doctors
may adjust dosages to account for individual factors such as genes. As another example, in
dynamic pricing, different values of customer rebates (e.g., from $10\%$ to $40\%$) can be
viewed as continuous {treatments} offered to the customer. The duration of or intensity of exposure to an intervention,
such as a job training program, can be considered as a continuous treatment as well.
Treating such variables as discrete, e.g., by discretizing the data, must rely on ad-hoc
modeling and may impede the fidelity of evaluation and the performance of optimized policies. 
Interpreting treatments as continuous is helpful even if not all continuum values are observed
since such an interpretation allows off-policy evaluation to learn
across treatments that are different but close.

We present a framework for policy evaluation and optimization with
continuous treatments. Our proposed estimator, introduced in Sec.~\ref{sec-methods},
effectively uses outcome data from data points where the treatment was close to the target
policy. In Sec.~\ref{sec-analysis}, we analyze the bias, variance, and
the mean-squared error of our policy evaluation. For the corresponding estimated optimal policy, we analyze the consistency of policy optimization in Sec.~\ref{sec-opt}. Finally, in Sec.~\ref{sec-experiments}, we conduct
experiments showing that our approach performs well. In particular, we consider
a case study with clinical data from a dataset of warfarin patients.

\section{Policy Evaluation Methodology}\label{sec-methods}
\vspace{-3pt}
\subsection{Problem Setup and Notation} \vspace{-7pt}
We study an observational dataset collected from interactions of the decision-making system assigning treatments to units $x_i$. For each interaction, the system observes an input (context or feature) $x_i \in \mathcal{X}$. Following the logging policy, the system assigns some treatment  $t_i \in \mathcal{T}$ with probability $Q_i$. Outcome data $y_i$ is observed which is generated from a joint distribution on the covariates, treatments, and induced outcomes, $(X,T,\{Y(t)\}_{t \in \mathcal{T}}) \sim \mathcal{D}$, unknown to us. The observational dataset comprises of $n$ i.i.d. observations of data $(x_i,t_i,y_i(t_i))) \sim \mathcal{D}$.


The term \textit{treatment} corresponds to \textit{arms} or \textit{actions} discussed in other works on off-policy evaluation. The generalized propensity score (GPS) is defined as $Q_i = f_{T\mid X}(T=t_i,X= x_i)= f_{T\mid X}(t_i\mid x_i)$ and generalizes the discrete propensity score $p_a = \mathbb{P}[t_i = a]$ for the continuous setting; we assume it exists \citep{hi04}. We assume the logging policy is known, which is reasonable when we have control over the system. Otherwise, the GPS can be imputed with standard approaches for predicting conditional densities such as regression under parametric noise models or kernel density estimation. 

Off-policy evaluation estimates the policy value $$\textstyle V_{\tau} = \mathbb{E}[Y(\tau(X))]$$ the value of the expected outcomes induced by the policy $\tau(X)$, corresponding to potential outcomes under the Neyman-Rubin causal framework. $Y(t)$ denotes the potential outcome of a unit had it received treatment $t$ \citep{r74}. We require standard assumptions in causal inference of \textit{unconfoundedness} (also known as ignorability) and \textit{common support}. Unconfoundedness asserts that $Y_i(t) \perp T_i \mid X_i$ for all $t$: treatment is exogenous and its assignment depends only on $x_i$. The data generation process described above is consistent with unconfoundedness. The \textit{common support} condition requires that $f_{T\mid X}(\tau(x_i)\mid x_i) > 0$: otherwise, if possible treatments may never be observed in the dataset, there is no chance for accurate estimation of their respective outcomes.

The task of policy optimization is to determine the optimal policy within a restricted function class of policies $\mathscr{T}$.  Since the optimal policy is deterministic (for each user, assign the optimal treatment), we focus on evaluating deterministic policies. The empirically optimal policy $\hat{\tau} \in \argmin_{\tau\in \mathscr{T}} \hat{v}^n_\tau$ is the policy minimizing the estimated value from our proposed estimator. The best-in-class policy $\tau^* \in \argmin_{\tau\in \mathscr{T}}  V_\tau$ minimizes the unknown expected policy value.
\subsection{Related Work}
\vspace{-7pt}
Previous approaches for off-policy evaluation broadly include the direct method (DM), inverse propensity weighted estimators (IPW), or estimators which combine them. The direct method estimates the relationship between outcomes and the union of covariates $x$ and treatment $t$, $r(X,T) = \mathbb{E}[Y\mid X,T]$, and generates a plug-in estimator. By unconfoundedness, regression-based estimation of the conditional mean function $\E[\E[Y|X,T=\tau(X)]]$ corresponds to estimation of the potential outcome \citep{r74}. However, this approach is subject to issues with model misspecification and without addressing dataset imbalance from a logging policy, may over or under-estimate the relevance of outcomes under a different policy \citep{bl09}. In \citep{kgdh2015}, the authors have investigated the evaluation of continuous treatment effects with the Super-Learner, an ensemble model which incorporates multiple models of the entire dose-response surface.


IPW-based estimation normalizes the observed outcomes by the inverse of the $\textit{propensity weights}$ of the logging policy  \citep{ht52}. IPW estimation corrects distribution mismatch by averaging outcomes over a new dataset created out of reweighted instances where the logging and target policies assign the same treatment \citep{lwlw11}. IPW is unbiased, with a slower rate of convergence dependent on the number of treatments \citep{bl09}, and it is optimal in the sense of minimax efficiency when no additional information about the reward structure is available \citep{wad2017}. However, dividing by the propensity score can inflate the variance of IPW estimators.

The doubly-robust (DR) estimator combines DM and IPW estimators. When the direct estimate of the reward estimator is biased, such as when using non-parametric or high-dimensional regression of $\mathbb{E}[Y\mid X,T]$, the doubly robust estimator weights the model residuals by inverse propensity weights in order to remove the bias. DR achieves a multiplicative bias when propensities are estimated and its convergence requires only that one of the estimators are consistent \citep{pb2016,dell2014}. Recent work in \citep{pb2016,wad2017} switches between using an IPS and reward estimator, using the reward estimator when the propensity is smaller than some threshold which optimizes the MSE bias-variance trade-off. In \citep{sj15,js15norm}, the authors propose counterfactual risk minimization for policy optimization by minimizing an upper bound on the MSE of an IPW-based estimator.

The continuous setting for policy evaluation and optimization presents new challenges. We note that the generalized propensity score as introduced in \citep{hi04} is analyzed in the context of treatment effect evaluation, and is used in practice to motivate appropriate discretizations of a continuous treatment variable for assessing balance. Policy optimization in discrete action spaces is generally reduced to a weighted multi-class classification problem, where the classes are treatments and are weighted by their off-policy evaluation. For each context, the policy determines the action which provides highest rewards as its classification label \citep{dell2014}. However, policy optimization in the continuous setting will be fundamentally different since the problem does not decompose into discrete classes of outcomes.

\subsection{Off-Policy Continuous Estimator}
\vspace{-7pt}
Previous IPW approaches for off-policy evaluation in discrete action spaces propose the following estimator, which filters the observational dataset by rejection and importance sampling:
\[\textstyle  \mathbb{E}\left[ \frac{1}{n} \sum_i \frac{y_i}{Q_i } \mathbb{I}\{\tau_i(x_i) = t_i\} \right] \]
In the continuous setting, we will not be able to employ rejection sampling since $\mathbb{P}[\tau(x_i)=t_i] = 0$ for any continuous probability density. The rejection sampling term $\mathbb{I}\{\tau(x_i) = t_i\}$ can be viewed as a Dirac delta function, $\delta_{\tau(x_i)}(t_i)$, and in the discrete case, it enforces that the only outcome data used for estimation are the observations where the same treatment was observed under the logging policy and is assigned by the target policy. For continuous treatments, our proposed estimator re-weights the dataset to consider outcomes where the observed treatment and off-policy treatment are close.

We propose the continuous-treatment off-policy evaluator, denoted as $ \hat{v}_{\tau}$, which smoothly relaxes the unit mass of the Dirac delta function using a kernel function $K(u)$:
\[  \hat{v}_{\tau} = \frac{1}{n h} \sum_{i=1}^n K\left( \frac{\tau(x_i) - t_i}{h} \right)\frac{y_i}{Q_i} \]
Properties of the kernel function $K(u)$ include symmetry about the origin ($\int u K(u) du = 0$) and that it integrates to 1 ($\int K(u) du = 1$). Kernel density estimates, also known as Parzen-window estimation, can be viewed as smooth nonparametric generalizations of computing histogram `buckets'. Instead of assuming a specific parametric statistical model, kernel density estimation assumes smoothness of the underlying joint density \citep{h09}. We state our results for univariate kernels where $\mathcal{T} \subseteq \mathbb{R}$ and note that analogous results hold if we use multidimensional kernel functions. When $\mathcal{T} \subseteq \mathbb{R}^d$, the estimator takes the form $\hat{v}_{\tau} = \frac{1}{n } \sum_{i=1}^n \vert H \vert^{-\frac{1}{2}}K(\vert H \vert^{-\frac{1}{2}}  (\tau(x_i) - t_i) )\frac{y_i}{Q_i}$, where $H$ denotes a bandwidth matrix. Examples of kernels include Gaussian kernels, where $K(u) = \frac{1}{\sqrt{2 \pi}} e^{-\frac{u^2}{2}}$ or the Epanechnikov kernel, $K(u) = \frac{3}{4}(1-u^2)\mathbb{I}\{\vert u \vert \leq 1 \}$.

This approach extends the IPW and rejection sampling approach taken in discrete treatment spaces to continuous treatment spaces. The extent of the kernel smoothing, parametrized by the bandwidth $h$, can be chosen to minimize the mean-squared-error (MSE). In particular, our estimator differs from using the direct method with kernel-based regression of the conditional density $f_{Y\mid T,X}$ and evaluation of the estimate at the treatment policy. Thus, we avoid the curse of dimensionality: kernel regression performs dramatically worse as the number of covariate dimensions increases, whereas the convergence rates of our method rely only on the treatment dimension \citep{pu99}.




In the special case that the logging policy is unknown such that propensities $Q_i$ must be imputed, and if a dose-response estimator $\hat{r}(t, x)$ is available, our estimator can be extended to a doubly robust one that has bias in excess of that in Thm.~\ref{thm-bias} that is multiplicative in the estimation biases of propensities and dose response:
\begin{align*}&\textstyle v_{\tau}^\text{DR} =\\&\textstyle \frac{1}{n} \sum_{i=1}^n \left[ \hat{r}(\tau(x_i), x_i) + \frac{1}{hQ_i}K\left(\frac{\tau(x_i) - t_i}{h}\right)(y_i - \hat{r}(t_i,x_i)) \right]  \end{align*}
\subsubsection{Self-Normalized Propensity Weight estimator}
\vspace{-5pt}
As discussed in \citep{js15norm}, IPW methods can suffer from variance in estimates due to the propensity weights. Normalizing the IPW estimator by $\sum \nicefrac{1}{Q_i} K(\frac{\tau(x_i) - t_i}{h})$ maintains consistency but can reduce variance by adjusting estimates of the treatment space that would have been be sampled with greater or lower probability.
\[ \textstyle  \hat{v}_{\tau}^{\text{norm}} = \frac{ \sum_{i=1}^n \frac{y_i}{Q_i} K\left( \frac{\tau(x_i) - t_i}{h} \right)}{\sum\frac{1}{Q_i} K\left(\frac{\tau(x_i)-t_i}{h}\right)} \]
\vspace{-10pt}
\subsubsection{Practical Concerns}
\vspace{-5pt}
When implementing our off-policy evaluation estimator in practice, some adjustments need to be made for empirical performance.

\textbf{Bandwidth selection}: Selecting a good bandwidth is key
to good evaluation and optimization. We compute the asymptotically optimal bandwidth in Thm.~\ref{thm-bias} below, but beyond the order in $n$, the expression includes constants that are generally unknown a priori. In the case of kernel density estimation, the literature focuses on methods for bandwidth selection which do not incorporate loss scalings and would perform poorly in our setting \citep{pm2009}.
Instead, we propose to select the optimal bandwidth via a plug-in estimator, estimating the quantities in the expression for optimal bandwidth (eq.~\ref{eq:opt-bandwidth}): we estimate the conditional density via kernel density estimation and subsequently estimate the second derivative and the conditional expectation via numerical integration.



\textbf{Boundary bias}: If the treatment space is bounded, the kernel may extend past the
boundaries where necessarily no data point exists, biasing boundary estimates downwards. This can be addressed by truncating and
normalizing the kernel: if $\mathcal T=[T_\text{lo},T_\text{hi}]$ then we
simply divide each term in our estimator by
$
\int_{T_\text{lo}}^{T_\text{hi}}K((\tau(x_i)-t)/h)dt
$. 
%

\textbf{Clipping propensity weights}: When using IPW-based estimators, in practice if the propensity score is very small, it is clipped with some threshold $\theta$, e.g., $0.1$. This introduces additional bias but may significantly reduce the variance, yielding
smaller total error.
\vspace{-5pt}
\section{Off-Policy Evaluation Analysis}\label{sec-analysis}
\vspace{-5pt}
\subsection{Bias and Variance of Kernelized Policy Evaluation} \vspace{-10pt}
We compute the bias and variance of the estimator $\hat{v}_\tau$ and prove consistency. Some technical assumptions are required for the analysis:
\begin{assumption}\label{biasasn1}
The conditional outcome and treatment densities, $f_{Y\mid T, X}$ and $f_{T\mid X}$, exist.\vspace{-3.5pt}
\end{assumption}
\begin{assumption}\label{biasasn2}
The conditional outcome density $f_{Y\mid T, X}$ is twice differentiable and the conditional treatment density $f_{T\mid X}$ is differentiable.\vspace{-2pt}
\end{assumption}

\begin{assumption}\label{biasasn3}
Outcomes $y_i$ are bounded with finite second moments.\vspace{-2pt}
\end{assumption}
\begin{assumption}\label{biasasn4}
Common support between the treatment propensities observed in the data and the treatment policy $\tau(x_i)$: $f_{T\mid X}(\tau(x_i),x)\geq a > 0$, for almost everywhere $x$ and some fixed $a$.\vspace{-2pt}
\begin{assumption}\label{biasasn5}
(Unconfoundedness) $Y_i(t) \perp T_i \mid X_i$ for all $t$: treatment is exogenous and its assignment depends only on $x_i$.\vspace{-2pt}
\end{assumption}
\end{assumption}


\begin{theorem}
\label{thm-bias}
Under Assumptions \ref{biasasn1}-\ref{biasasn5},
the bias and variance of $\hat{v}_{\tau}$ are:
\begin{align*}& \textstyle \Bias(\tau) \textstyle:=  \mathbb{E}\left[\hat{v}_{\tau} - V_\tau   \right] = \\&   \kappa_2(K)  \mathbb{E}\left[ \int  \frac{y_i}{2} \frac{\partial^2}{\partial T^2} f_{Y\mid T, x}(y_i, \tau(x_i))  dy \right]{h^2} + {o(h^2)}\\
&\textstyle \Var({\tau}) \textstyle:= \mathbb{E}[(\hat{v}_\tau - \mathbb{E}[\hat{v}_\tau] )^2] =  \\& R(K) \mathbb{E}\left[\frac{\mathbb{E}[Y^2 \mid \tau(X), X]}{f_{T \mid X}(\tau(X), X)}\right]\frac1{nh}  + O(\frac{h^4 }{n}) + o\left(\frac{1}{nh}\right)
\end{align*}
where $\kappa_2(K)=\int u^2K(u)du$ and $R(K)=\int K(u)^2 du$ are the second moment and roughness of the kernel, respectively.
\end{theorem}
\vspace{-10pt}
\begin{proof}[Proof outline]
The theorem follows by applying Bayes' rule with the GPS and Taylor expansion of $f_{Y\mid T, X}$ around $\tau(x_i)$. Details in Appendix~\ref{pf-bias}.
\end{proof}
The bias introduced by kernel density estimation is $O(h^2)$ and depends on the curvature of the unknown density $f_{Y \mid T,X}$ evaluated at the policy $\tau(X_i)$: if the outcome distribution changes rapidly with small changes in treatment value,  our approach for leveraging local information will incur more bias. The variance depends inversely on the generalized propensity score. As expected, the estimator may have high variance in regions where we are unlikely to observe treatment $\tau(x_i)$.
\subsection{Mean Squared Error and Consistency}
\vspace{-7pt}
We analyze mean squared error derived from bias and variance and characterize the optimal bandwidth. Intuitively, the bandwidth controls the scale of proximity we require on treatments: a bandwidth too large introduces high bias because we simply average over the entire dataset, while small bandwidths increase variance.

\begin{theorem}\label{thm-mse}
Under Assumptions \ref{biasasn1}-\ref{biasasn5}, the MSE of $\hat{v}_{\tau}$ is
\begin{align*}\textstyle
&\textstyle \mathbb{E}\left[(\hat{v}_{\tau} - V_\tau)^2   \right]=\\&
\textstyle \frac{R(K)}{nh} \mathbb{E}\left[\frac{\mathbb{E}[Y^2 \mid \tau(X), X]}{f_{T \mid X}(\tau(X) \mid X)}\right]+ O(h^4) + O\left( \frac{h^4 }{n} \right) + o\left(\frac{1}{nh}\right)  \end{align*}
and the optimal bandwidth $h^*$ is $\Theta(n^{-\frac{1}{5}})$.
\[ \textstyle h^* =\left(\frac{{R(K)} \mathbb{E}\left[\nicefrac{\mathbb{E}[Y^2 \mid \tau(X), X]}{f_{T \mid X}(\tau(X), X)}\right]}{4 (\mathbb{E}\left[ \int  \frac{y_i}{2} \frac{\partial}{\partial T^2} f_{Y\mid T, x}(y_i, \tau(x_i))  \kappa_2(K) dy \right])^2 n}\right)^{\frac{1}{5}}\]
\end{theorem}   \vspace{-12pt}    \label{eq:opt-bandwidth}\begin{proof}[Proof outline]
The theorem follows from the bias-variance decomposition of MSE and using Theorem \ref{thm-bias}, then optimizing over $h$. (Appendix~\ref{pf-mse})
\end{proof}
\begin{theorem}\label{thm-consistency}
Under Assumptions \ref{biasasn1}-\ref{biasasn5}, if $1/(nh)\to0$ then $\hat{v}_{\tau}$ is consistent for $V_\tau$:
\[\textstyle \frac{1}{n h} \sum_{i=1}^n K\left( \frac{\tau(x_i) - t_i}{h} \right)\frac{y_i}{Q_i} \to_p  V_\tau\]

\end{theorem}
\vspace{-10pt}
\begin{proof}[Proof Outline]
Follows from convergent MSE and Markov's inequality. Full proof in Appendix~\ref{pf-consistency}
\end{proof}


\begin{corollary}\label{thm-normalized-consistency}
Under Assumptions \ref{biasasn1}-\ref{biasasn5}, if $1/(nh)\to0$ then
the self-normalized off-policy evaluation estimator is consistent for $V_\tau$.
\vspace{-3pt}
\[ \textstyle\hat{v}_{\tau}^{\mathrm{norm}} = \frac{ \sum_{i=1}^n \frac{y_i}{Q_i} K\left( \frac{\tau(x_i) - t_i}{h} \right)}{\sum\frac{1}{Q_i} K\left(\frac{\tau(x_i)-t_i}{h}\right)} \to_p V_\tau \]
\vspace{-7pt}
\end{corollary}
\vspace{-7pt}
\begin{proof}
The result follows from Slutsky's theorem since $\frac{1}{nh}\sum_{i=1}^n\frac{1}{Q_i} K\left(\frac{\tau(x_i)-t_i}{h}\right) \to_p 1$.
\end{proof}
\vspace{-7pt}
\section{Continuous Policy Optimization}\label{sec-opt}
\vspace{-3pt}
Accurate off-policy evaluation is a necessary prerequisite for policy optimization, the task of estimating which treatment policy minimizes expected desired outcomes. We analyze how the empirically optimal policy, the policy minimizing the off-policy evaluations, performs out-of-sample.

For a constrained policy class, such as a space of linear policies ($\mathscr{T} = \{ \tau(x) =  \beta^\intercal x : \norm{\beta}_2 \leq W_2 \}$), the policy optimization problem can be interpreted as a weighted empirical risk minimization problem over a constrained policy space where we find $\hat{\tau} \in \underset{\tau\in \mathscr{T}}{\argmin} \frac{1}{n h} \sum_{i=1}^n K\left( \frac{\tau(x_i) - t_i}{h} \right)\frac{y_i}{Q_i} $.
Gradients can be computed easily with respect to the kernel function, applying the chain rule to $\tau(x)$, and we provide additional examples in the Appendix. Equivalently we can optimize the other estimators $\hat{v}_\tau^{\text{norm}}, \hat{v}_{\tau}^{\text{DR}}$ and incorporate variance regularization.  The nonconvex optimization can be solved by a numerical optimizer such as L-BFGS or gradient descent, but generally with no guarantees for global convergence. In practice we take the best solution from random restarts. 

\vspace{-7pt}
\subsection{Consistency of Policy Optimization}
\vspace{-7pt}
Our analysis bounds the generalization error for the empirical risk-minimizing policy, the error incurred by minimizing the empirical risk instead of the unknown expected risk.  Generalization bounds for this problem follow from \citep[Thm.~8]{bm2002}, and depend on the Rademacher complexity of the loss function class. The empirical Rademacher and Rademacher complexity of a function class $\mathscr{T}$ are, respectively, defined as:
\begin{align*}\textstyle \hat{\mathcal{R}}_n(\mathscr{T}) =& \textstyle\mathbb{E}\left[ \sup_{f \in \mathscr{T}} \left| \frac{2}{n} \sum_{i=1}^n \sigma_i f(x_i) \right| \mid x_1 , ... x_n  \right]\\\textstyle \quad
{\mathcal{R}}_n(\mathscr{T}) =& \mathbb{E}[\hat{\mathcal{R}}_n(\mathscr{T})]\end{align*}
where $\sigma_i$ are iid Rademacher random variables, symmetrically $1$ or $-1$ with probability $\frac{1}{2}$. Restricting the function class provides better generalization error by reducing the Rademacher complexity: a function class which is less able to fit arbitrary data sequences is less vulnerable to overfitting.

\begin{assumption}\label{consasn5}
Outcome values $y_i$ are bounded on the interval $[-\bar{M}_y, \bar{M}_y]$. The inverse propensity weight,$\frac{1}{Q_i}$, is bounded on $[1,\bar{M}_Q]$.
\end{assumption}

\begin{assumption}\label{consasn6}
The kernel function $K(u)$ is bounded by $\bar M_K$ and has Lipschitz constant $L_K$.
\end{assumption}

\begin{theorem}

Under Assumptions \ref{biasasn1}-\ref{consasn6}, for any integer $n$ number of samples and any $0 < \delta < 1$, with probability at least $ 1- \delta$, every $\tau \in \mathscr{T}$ satisfies:
\begin{align*}&\left|\textstyle V_\tau - \hat v_{\tau}\right|  \leq \\&
\textstyle \frac{L_K\bar{M}_Q\bar M_y}{h^2} \mathcal R_n(\mathscr{T}) + \frac{\bar{M}_y\bar{M}_Q\bar M_K}{h}\sqrt{\frac{2 \log(2/\delta)}{n}} + \left|\Bias(\tau)\right| \end{align*}

\end{theorem}\vspace{-5pt}
\begin{proof}[Proof Outline.]
The result follows from the Rademacher generalization bound \citep[Thm.~8]{bm2002}
concentrating $\E\hat v_\tau$ near $\E\hat v_\tau$, Thm.~\ref{thm-bias} relating $V_\tau$ to $\E\hat v_\tau$, and the Rademacher comparison lemma \citep[Thm.~4.12]{ledoux1991probability}. The full proof is provided in Appendix~\ref{pf-opt}.
\end{proof}
\vspace{-5pt}
\begin{corollary}
Let $\hat\tau\in\argmin_{\tau\in\mathscr T}\hat v_\tau$ and $\tau^*\in\argmin_{\tau\in\mathscr T}V_\tau$. Then, under Assumptions \ref{biasasn1}-\ref{consasn6}, with high probability the regret of $\hat\tau$ satisfies
$$\textstyle
V_{\hat\tau} - V_{\tau^*} \leq O_p(\frac1{h^2}\mathcal R_n(\mathscr T)+\frac{1}{h\sqrt{n}}+h^2)
$$
\end{corollary}
\vspace{-10pt}
The corollary shows that the regret of our policy optimizer converges
to zero, i.e., achieves best-in-class performance,
as long as $h=o(1)$, $h=\omega(\sqrt{\mathcal R_n(\mathscr T)})$, and
$h=\omega(1/\sqrt{n})$.
%
%
As an example, consider a function class of linear decision rules with bounded norm:
$\mathscr T=\{\tau(x)=\beta^\intercal x:\norm{\beta}_2 \leq W_2\}$.
Assuming that $\norm{X_i}_2 \leq X_2$, \cite{kst2009} shows that
the Rademacher complexity of this class is bounded as $ \hat{\mathcal{R}}_n(\mathscr T) \leq \frac{W_2 X_2}{\sqrt{n}}$.
%
Therefore, the optimal bandwidth $h = \Theta(n^{-\frac{1}{5}})$ ensures consistent
policy learning of the best linear policy. Similar results hold for policies in a
bounded ball of a reproducing kernel Hilbert space.

\textbf{Variance regularization}: If the optimization space includes a policy which assigns treatments arbitrarily far from the observed treatments, such a policy trivially minimizes the loss by forcing $K(\frac{\tau(x_i) - T_i}{h}) \to 0$. Regularizing the objective by the estimated sample standard deviation $\frac1n \sqrt{\sum_{i=1}^n (K( \frac{\tau(x_i) - t_i}{h} )\frac{y_i}{Q_i} - \hat{v}_\tau)^2}$ of the policy evaluation should mitigate this effect from overly expressive policy classes.\vspace{-5pt}
\section{Experiments}\label{sec-experiments}
\vspace{-3pt}
\subsection{Validation on Synthetic Data}\label{sec-synthetic}
\vspace{-7pt}
We first consider a controlled setting with synthetic data to illustrate our method.
We consider $y = 2\vert x - t\vert^{1.5} + 0.2\epsilon$, where $\epsilon \sim N(0,1)$ and $x \sim \Unif[0,1]$.
We consider treatment assignment that is either completely randomized
uniformly on the interval $[-0.5,1.3]$ without regard to $x$ or
treatment assignment that is confounded by $x$ and is normally distributed as
$T \mid X \sim N(x + 0.1,0.5 )$.
The optimal policy is linear and sets $\tau(x_i) = \beta x_i$ where $\beta = 1$.

We consider how the performance of off-policy evaluation changes with $n$ by evaluating the optimal
policy with $n$ observational data points generated either using completely randomized
treatments or treatments confounded by $x$, clipping generalized propensities below 0.1.
We use the Epanechnikov kernel with the self-normalized estimator and estimate the bandwidth by using kernel density estimation of $f_{Y,T,X}$ and $f_{T,X}$ for the conditional density $f_{Y \mid T,X} = \frac{f_{Y,T,X}}{f_{T,X}}$. From this estimate we obtain an estimate for the second derivative $\frac{\partial^2}{\partial T^2}f_{Y \mid T,X} $ by numerical differentiation and compute an approximate conditional expectation of $\mathbb{E}[Y_i^2 \mid \tau(x_i), X]$ via numerical integration. Since computing the bandwidth is numerically intensive, we compute it for one value of $n_0$ and adjust it for different $n_i$ by multiplying by  $(\frac{n_0}{n_i})^{\nicefrac{1}{5}}$. We compare to standard clipped-IPW discrete-treatment policy evaluation by discretizing the treatments into 10 evenly sized bins from the minimal to maximal observed treatment, computing the discrete propensity score by integrating the GPS over the bin (``discretized OPE''). We also compare against the direct method, using either a trained random forest regressor (``DM RF'') or polynomial regression of order 3 (``DM Poly''). 

\begin{figure}[t!]
\RawFloats
\centering
\captionsetup{width=0.9\textwidth}
\centering\includegraphics[width=\textwidth]{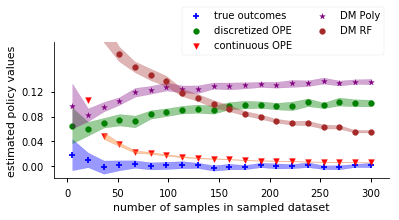}
\small\caption{Off-policy evaluations and 95\% confidence intervals as $n$ increases, evaluating a linear treatment policy where $\beta= 1$ with normal sampling (model in Sec.~\ref{sec-synthetic}).}
\label{fig:consistency_normal}
\includegraphics[width=\textwidth]{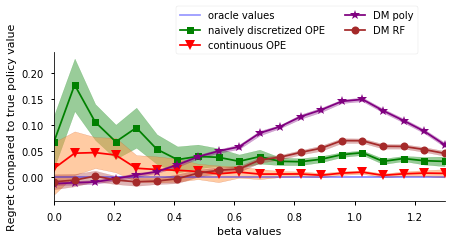} \small\caption{Regret of evaluations of a linear treatment policy compared to the true value, n=300, over different values of $\beta$ with completely randomized sampling. }
\label{fig:evaluation-normal}

\end{figure}

For each $n$ between 10 and 300,
we simulate 50 replications of the process. The results in Fig.~\ref{fig:consistency_normal} include 95\% confidence intervals around the mean over replications.
In both settings our policy evaluation indeed converges to the truth and
while the discretization approach performs reasonably well, it is systematically
biased, inconsistent, and has larger variance. The discretization is sensitive to the distribution of the data and the variation in the unknown true relationship between covariates, treatment, and outcome. In practical settings with real data, it is unclear what the best discretization would be. 

\begin{figure}[t!]
\RawFloats
\centering\includegraphics[width=\textwidth]{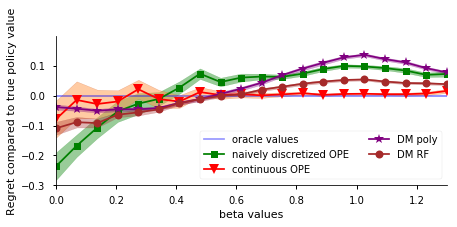}\small\caption{Regret of evaluations of a linear treatment policy compared to the true value, n=300, over different values of $\beta$ with confounded sampling. }\label{fig:linear_treatment_n_300}
\centering
\includegraphics[width=\textwidth]{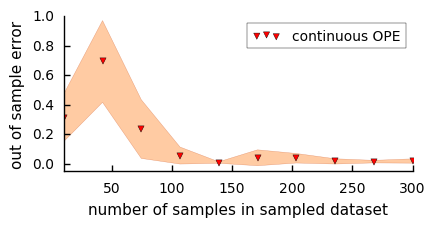}\small\caption{Out-of-sample error of empirically optimal policy from off-policy evaluation as $n$ increases.}
\label{fig:pol-opt-consistency}
\vspace{-7pt}
\end{figure}

To consider off-policy optimization in this simple setting, we fix $n=300$ and
evaluate linear policies with $\beta$ ranging over $[0,1.3]$.
Again we consider 50 replications. Figs.~\ref{fig:evaluation-normal} and~\ref{fig:linear_treatment_n_300}
show 95\% confidence intervals around the mean over replications.
Our policy evaluations are tight near the optimum and subsequent optimization over the range of $\beta$ values is consistent with the true optimal $\beta$. In Fig.~\ref{fig:pol-opt-consistency} we evaluate the consistency of policy optimization by analyzing the out-of-sample error of the empirically optimal policy computed on datasets of $n$ varying from 10 to 300. We compute 20 replications and display 95\% confidence intervals around the mean, observing that following the theory, the out-of-sample error from off-policy evaluation converges to zero as n increases.


\vspace{-7pt}

\subsection{Policy Optimization Simulation }\label{sec-high-d}
We consider a similar controlled setting in higher dimension with richer dose-response structure, still with synthetic data, and illustrate the resulting outcome distribution under various learned policies. We randomly generate independent ten-dimensional covariates ($d=10$), normally distributed with zero mean and randomly generated covariances (following a normal distribution which is offset for positivity). The true outcome model is quadratic in $T$: we set the noiseless outcome as $y_i = \beta_T^\intercal T \beta_x^\intercal x_i + \beta_{x,T}^\intercal x_i T_i + (T_i - \beta_{x,T^2}^\intercal x)^2$ where $\beta_x \sim N(0, I_{d}) $, $\beta_{x,T}\sim N(0, 1.5 I_{d})$, and $\beta_{x,T^2}\sim N(0, I_d)$. We induce sparsity on the coefficients by independently and randomly sampling 3 covariates to remain positive on each coefficient vector $\beta_x$, $\beta_{x,T}$, and $\beta_{x,T^2}$. We include a constant treatment effect interaction term of $\beta_T =-5$.  

We sample treatments as normally distributed conditional on covariates, $T\sim N(\theta^T x,4)+2x_1 +4x_2 -3x_3$ and generate a training dataset of 400 instances $(x_i,y_i,T_i)$ and evaluate on a test set of 1000 instances. Outcomes are generated from the model $y_i$ with additional i.i.d. mean-zero Gaussian noise with variance $5$.
For policy learning, we consider the case that the propensity model is well-specified but unknown and impute the generalized propensity score from the training data via linear regression. 
\begin{figure}[t!]
	\includegraphics[width=.95\textwidth]{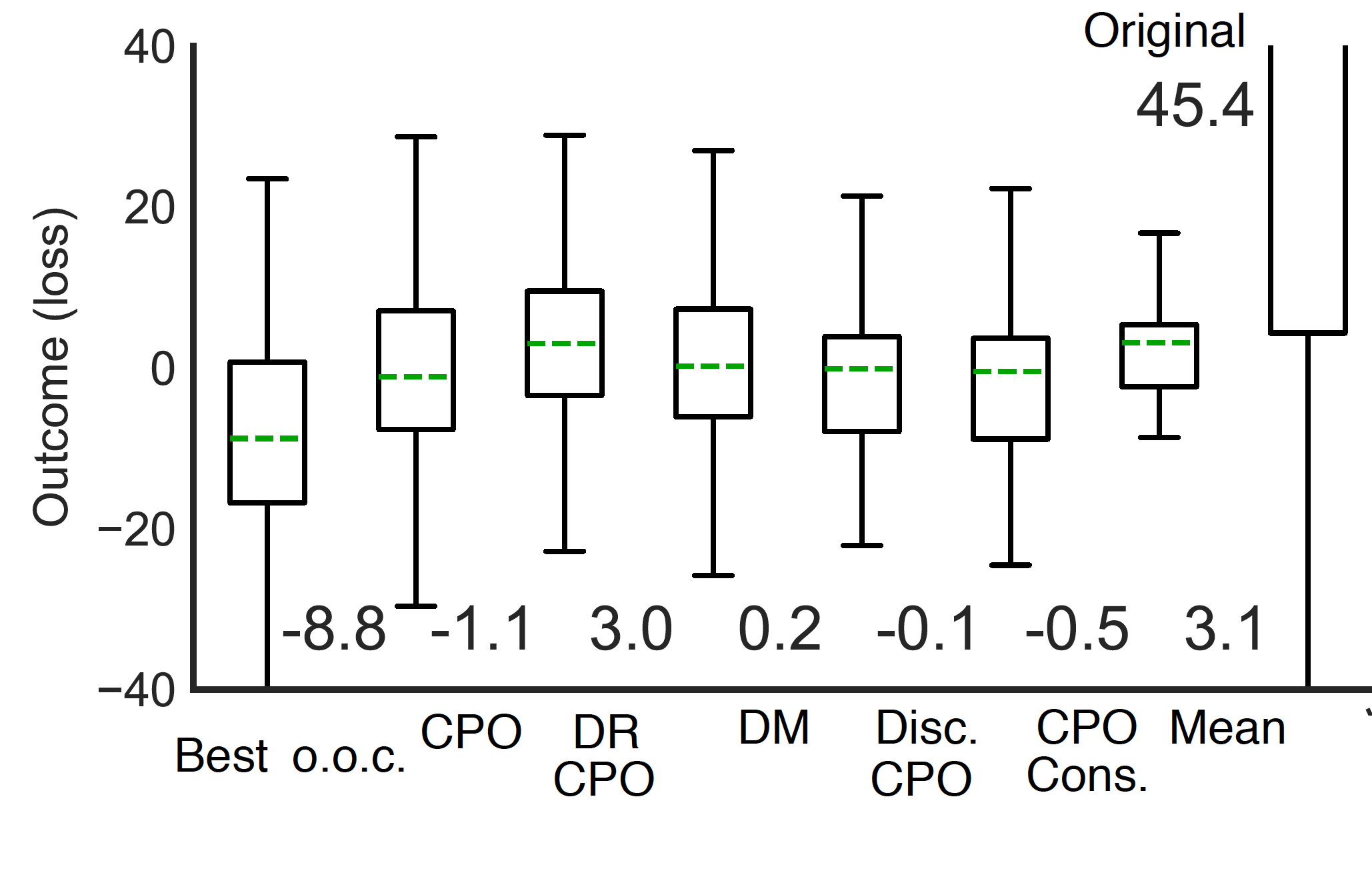}
	{\caption{Boxplot of the distributions of outcomes for policy learning on 10-dimensional covariates with a quadratic outcome model (Sec.~\ref{sec-high-d}), displaying mean loss values.}\label{fig:simulated_high_d}}
	\vspace{-10pt}
\end{figure}

In Fig.~\ref{fig:simulated_high_d} we compare the outcome distributions under learned policies using a box plot, displaying the means on the right. For reference we compute the best treatment assignment for each $x_i$ given the full counterfactual model (best out-of-class (o.o.c)). We evaluate continuous off-policy evaluation over linear policies with a bandwidth of 2.6. When optimizing the off-policy estimator for the linear policy, we use the L-BFGS algorithm with multiple random restarts since the objective function is non-convex and L-BFGS performs well even in the non-convex setting \cite{l99}. We evaluate the direct method (DM), using a random forest regressor with a linear policy space, which we optimize by using the numerical differentiation available with BFGS. We use the same random forest regressor for doubly-robust continuous policy optimization (DR CPO). We also consider a discretization approach which optimizes over a continuous treatment policy (linear) but evaluates performance by discretizing treatment into $10$ uniformly sized bins, running standard self-normalized CPE. Discretizing the resulting linear policy yields a constant policy in this setting: we compare to the best constant policy found using continuous policy evaluation (CPE, cons.). Finally, the baseline is a constant policy which assigns the mean dose. 
Comparing the results, we see that off policy evaluation is able to improve upon the mean risk of other methods and nears the performance of the best treatment assignment with full information (which is out of the linear policy class). While the best constant policy found using OPE has good performance in the sense of mean risk, the linear policy is better able to personalize treatment based on covariates.

\subsection{Warfarin case study}\label{sec:warfarin}
\vspace{-7pt}
Unlike for discrete off-policy evaluation, no evaluation datasets are available for continuous treatments with full counterfactuals. We evaluate our estimator in an experimental setting by developing a case study from a PharmGKB \cite{iwpc09} dataset on warfarin dosing which includes information on patient covariates, final therapeutic dosages, and patient outcomes (INR, International Normalized Ratio). 
Warfarin is a blood thinner whose therapeutic dosage varies widely across patients and whose administration must be closely monitored to prevent adverse side effects. Previous work on predicting dosage policies
\citep{bb15,k16} has evaluated accuracy based on prediction of the correct category of dosage, ``low'' (<21 mg/wk), ``medium'' (> 21 mg/wk,< 49 mg/wk) or ``high'' (> 49 mg/wk). However, clinical guidelines suggest fine adjustments to dosage (15-20\%) during monitoring, and recommend splitting tablets to deliver precise treatment \cite{iwpc09}. Therefore, treating warfarin dosage as a continuous variable better captures the inherently continuous nature of dosage amounts.

We develop a semi-simulated study by simulating a dosage process in a way that allows us to simulate counterfactual outcomes.  Following the procedure set out in \cite{iwpc09}, we consider correct prediction as being within 10\% of the therapeutic dose $T_i^*$, since measurements of patient INR are inherently noisy and dose is adjusted until the patient INR presents within a target range. Since the clinical risk of incorrect dosage increases with absolute distance from the target range \cite{clinicalrisk11}, we use a semi-simulated loss function of absolute distance from $[.9T_i^*, 1.1T_i^*]$, instead of simulating unavailable INR outcomes: 
\[ \textstyle y(\tau(x_i)) = \max(\vert\tau(x_i) - T_i^*\vert-0.1T_i^*,0) \]
We sample our dosage data $T_i$ as a mixture of a patient's BMI z-score $Z_{\mathrm{BMI}} = \frac{x_{\mathrm{BMI}}- \mu_{\mathrm{BMI}}}{\sigma_{\mathrm{BMI}}}$ and i.i.d standard normal noise $\epsilon_i$, scaled to preserve the moments of the therapeutic dose distribution, $\mu_{T}^*, \sigma_T^*$, such that with $\theta = 0.5$:
 $\textstyle T_i = \mu_{T}^* + \sigma_T^* \sqrt{\theta} Z_{\mathrm{BMI}} + \sigma_{T}^*\sqrt{(1 - \theta)}  \epsilon$. 
  It follows that the propensity score is $ \textstyle f_{T\mid X}(T_i'=t \mid x_{\mathrm{BMI}}) = f_Z\left( \frac{t - \mu_T^* - \sigma_T^* \sqrt{\theta}Z_{\mathrm{BMI}}}{\sqrt{1 - \theta}} \right)$
where $\epsilon \sim N(0,1)$ and $f_Z$ is the continuous density of a standard normal random variable.


We impose bounds on the coefficient, $\beta_d \in [-\frac{T_{\mathrm{max}}}{.25d \mu_{X_d}}, \frac{T_{\mathrm{max}}}{.25d\mu_{X_d}} ]$, to prevent evaluating a policy with no overlap with the observed dataset, where $T_{\mathrm{max}}$ is the maximal treatment, $\mu_{X_d}$ is the mean of the $d$th covariate and $d$ denotes dimension. We run a priori feature selection on the full dataset before evaluating policy optimization, using the importance weights from a random forest regressor on the therapeutic dose to select the 81 most important features.

\begin{figure}[t!]
\includegraphics[width=\textwidth]{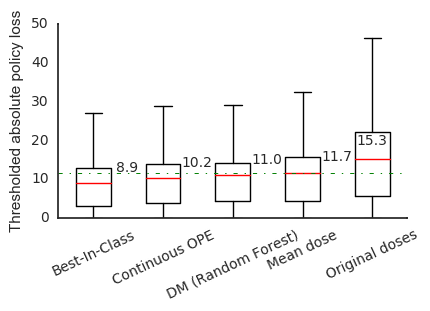}
{\caption{Boxplot of the distributions of thresholded absolute distances between policy and therapeutic doses, including mean loss values, for Warfarin study in Sec.~\ref{sec:warfarin}.}\label{fig:warfarin-comparison}}
\end{figure}
\begin{figure}[t!]
			\caption{Table of summary statistics of differences between estimated policy and true therapeutic doses}\label{tbl:warfarin_results}
	\begin{tabular}{@{}llll@{}}
		\toprule
		Policy & Mean L1 & Std. dev. L1 & Mean L2 \\ \midrule
		Best & 8.93 & 8.64 & 154.37 \\
		Cont. OPE & 10.19 & 10.19 & 207.78 \\
		DM & 11.02 & 10.96 & 241.68 \\
		Mean dose & 11.67 & 10.52 & 246.80 \\
		Original & 15.27 & 13.08 & 404.28 \\ \bottomrule
	\end{tabular}
\end{figure}
\vspace{-5pt}

We conduct policy optimization on these simulated outcomes and evaluate how the empirically optimal policy $\hat{\tau}_n$ performs on the thresholded loss function with absolute and squared penalties. The best-in-class linear treatment policy from median regression, $\tau^* \in \argmin_{\tau}  \mathbb{E}[\vert\tau(x_i)- T_i^*\vert \mid X, T]$, has access to information about the true therapeutic dose (``Best-in-class'' on the figure). We also evaluate the best linear model from a random forest regressor (DM estimator) for $\hat{r}(t_i, x_i)$. We compare against the linear policy found using our CPO method (``Continuous OPE'') which achieves a mean loss of 10.2. The baseline is a constant policy corresponding to the mean dose and for reference we plot the distribution of outcomes according to the original initial treatment assignment observed in the dataset, which doctors adjusted until a therapeutic dose was reached when patient INR was within the target range. We tested discrete off-policy optimization (POEM and NORM-POEM) with various uniformly-spaced discretizations or quantile-based discretizations of dosages, but the propensity scores are mostly zero or one and hamper the resulting optimization, illustrating the difficulty of finding appropriate discretizations for real datasets \cite{sj15,js15norm}.

Comparing the results in Fig.~\ref{fig:warfarin-comparison}, we see that our method is competitive with the best-in-class linear policy and improves upon the direct method, further reducing the median (Mean L1) and mean of the difference between the policy dose and therapeutic dose from the naive benchmark policy giving the mean dose. In Table~\ref{tbl:warfarin_results} we also report the squared distance from $[0.9T_i^*, 1.1T_i^*]$ (mean L2): we see that performance of the DM policy shows less improvement from giving the mean dose when we weight outliers more heavily, and results in larger variance in the distribution of absolute losses (std. dev L1). Our approach for policy optimization based on continuous off-policy evaluation works well in simulated and semi-experimental settings.
\vspace{-5pt}
\section{Conclusion}\vspace{-5pt}
We developed an inverse-propensity-weighted estimator for off-policy evaluation and learning with continuous treatments, extending previous methods which have only considered discrete actions. The estimator replaces the rejection sampling used in IPW-based estimators with a kernel function to incorporate local information about similar treatments. Our generalization bound for policy optimization shows that the empirically optimal treatment policy computed by minimizing the off-policy evaluation also converges to the policy minimizing the expected loss. We demonstrate the efficacy of our approach for estimation and evaluation on simulated data, as well as on a real-world dataset of Warfarin dosages for patients.

\textbf{Acknowledgments}

This material is based upon work supported by the National Science Foundation under Grant No. 1656996. Angela Zhou is supported  through the National Defense Science \& Engineering Graduate Fellowship (NDSEG) Program.


\small

\begin{thebibliography}{10}

\bibitem{bm2002}
Peter Bartlett and Shahar Mendelson.
\newblock Rademacher and gaussian complexities: Risk bounds and structural
  results.
\newblock {\em Journal of Machine Learning Research}, 2002.

\bibitem{bb15}
Hamsa Bastani and Mohsen Bayati.
\newblock Online decision-making with high-dimensional covariates.
\newblock {\em Management Science}, 2015.

\bibitem{bl09}
Alina Beygelzimer and John Langford.
\newblock The offset tree for learning with partial labels.
\newblock {\em Proceedings of the 15th ACM SIGKDD international conference on
  Knowledge discovery and data mining}, 2009.

\bibitem{dell2014}
Miroslav Dudik, Dumitru Erhan, John Langford, and Lihong Li.
\newblock Doubly robust policy evaluation and optimization.
\newblock {\em Statistical Science}, 2014.

\bibitem{clinicalrisk11}
Valentin Fuster, Lars~E. Ryden, Davis~S. Cannom, Harry~J. Crijns, Anne~B.
  Curtis, Kenneth~A. Ellenbogen, Jonathan~L. Halperin, G.~Neal Kay,
  Jean-Yves~Le Huezey, James~E. Lowe, S.~Bertil Olsson, Eric~N. Prystowsky,
  Juan~Luis Tamargo, and L.~Samuel Wann.
\newblock 2011 accf/aha/hrs focused updates incorporated into the acc/aha/esc
  2006 guidelines for the management of patients with atrial fibrillation.
\newblock {\em Circulation}, 2011.

\bibitem{h09}
Bruce Hansen.
\newblock Lecture notes on nonparametrics.
\newblock Technical report, University of Wisconsin, 2009.

\bibitem{hi04}
Keisuke Hirano and Guido Imbens.
\newblock {\em The Propensity Score with Continuous Treatments, in Applied
  Bayesian Modeling and Causal Inference from Incomplete-Data Perspectives: An
  Essential Journey with Donald Rubin's Statistical Family}, chapter~7.
\newblock John Wiley \& Sons, Ltd, 2004.

\bibitem{ht52}
Daniel Horvitz and Donovan Thompson.
\newblock A generalization of sampling without replacement froma finite
  universe.
\newblock {\em Journal of the American Statistical Association}, 1952.

\bibitem{iwpc09}
T~E International Warfarin Pharmacogenetics~Consortium, Klein, R~B Altman,
  N~Eriksson, B~F Gage, S~E Kimmel, M-T~M Lee, N~A Limdi, D~Page, D~M Roden,
  M~J Wagner, M~D Caldwell, and Johnson~J A.
\newblock Estimation of the warfarin dose with clnical and pharmacogenetic
  data.
\newblock {\em The New England Journal of Medicine}, 2009.

\bibitem{kst2009}
Sham Kakade, Karthik Sridharan, and Ambuj Tewari.
\newblock On the complexity of linear predicttion: Risk bounds, margin bounds,
  and regularization.
\newblock {\em Advances in Neural Information Processing Systems}, 2009.

\bibitem{k16}
Nathan Kallus.
\newblock Recursive partitioning for personalization using observation data.
\newblock {\em Proceedings of the Thirty-fourth International Conference on
  Machine Learning}, 2017.

\bibitem{kgdh2015}
Noemi Kreif, Richard Grieve, Ivan Dia, and David Harrison.
\newblock Evaluation of the effect of a continuous treatment: A machine
  learning approach with an application to treatment for traumatic brain
  injury.
\newblock {\em Health Economics}, 2015.

\bibitem{cccp09}
Gert~R. Lanckriet and Bharath~K. Sriperumbudur.
\newblock On the convergence of the concave-convex procedure.
\newblock {\em Advances in Neural Information Processing Systems 22}, 2009.

\bibitem{ledoux1991probability}
Michel Ledoux and Michel Talagrand.
\newblock {\em Probability in Banach Spaces: isoperimetry and processes}.
\newblock Springer, 1991.

\bibitem{l99}
Dong-Hui Li and Masao Fukushima.
\newblock On the global convergence of bfgs method for nonconvex unconstrained
  optimization problems.
\newblock {\em SIAM Journal on Optimization}, 2000.

\bibitem{lwlw11}
Lihong Li, Wei Chu, John Langford, and Xuanhui Wang.
\newblock Unbiased offline evaluation of contextual-bandit-based news article
  recommendation algorithms.
\newblock {\em Proceedings of the fourth ACM international conference on web
  search and data mining}, 2011.

\bibitem{pu99}
Adrian Pagan and Aman Ullah.
\newblock {\em Nonparametric Econometrics}.
\newblock Cambridge University Press, 1999.

\bibitem{pm2009}
Byeong Park and J.S. Marron.
\newblock Comparison of data-driven bandwidth selectors.
\newblock {\em Journal of the American Statistical Association}, 2009.

\bibitem{r74}
Donald Rubin.
\newblock Estimating causal eeffect of treatments in randomized and
  nonrandomized studies.
\newblock {\em Journal of Educational Psychology}, 1974.

\bibitem{s78}
Bernard Silverman.
\newblock Weak and strong uniform consistency of the kernel estimate of a
  density and its derivatives.
\newblock {\em The Annals of Statistics}, 1978.

\bibitem{sj15}
Adith Swaminathan and Thorsten Joachims.
\newblock Counterfactual risk minimization.
\newblock {\em Journal of Machine Learning Research}, 2015.

\bibitem{js15norm}
Adith Swaminathan and Thorsten Joachims.
\newblock The self-normalized estimator for counterfactual learning.
\newblock {\em Proceedings of NIPS}, 2015.

\bibitem{pb2016}
Philip Thomas and Emma Brunskill.
\newblock Data-efficient off-policy policy evaluation for reinforcement
  learning.
\newblock {\em Journal of Machine Learning Research}, 2016.

\bibitem{wad2017}
Yu-Xiang Wang, Alekh Agarwal, and Miroslav Dudik.
\newblock Optimal and adaptive off-policy evaluation in contextual bandits.
\newblock {\em Proceedings of Neural Information Processing Systems 2017},
  2017.

\end{thebibliography}

\bibliographystyle{plain}
\newpage
\onecolumn
\section{Appendix}

\normalsize
\subsection{Bias and Variance: Proof of Theorem \ref{thm-bias}}\label{pf-bias}
\begin{proof}

We provide full computations for the bias $\mathbb{E}\left[\hat{v}_{\tau} - V_\tau   \right]$ and variance $ \mathbb{E}[(\hat{v}_\tau - \mathbb{E}[\hat{v}_\tau] )^2]$

We compute the expectation of the estimator at one data point, $\frac{1}{h}K\left( \frac{\tau(x_i) - t_i}{h} \right)\frac{y_i}{Q_i} $, omitting the $\frac{1}{n}$ term. By linearity of expectation:
$$\mathbb{E}\left[\frac{1}{h}K\left( \frac{\tau(x_i) - t_i}{h} \right)\frac{y_i}{Q_i}\right] = \mathbb{E}\left[\sum_{i =1}^n \frac{1}{nh}K\left( \frac{\tau(x_i) - t_i}{h} \right)\frac{y_i}{Q_i}\right]$$

The analysis follows the structure of standard bias and variance calculations for kernel density estimation \cite{pu99}. We can express the conditional expectation of the kernel estimator via the integral convolution of the kernel and the conditional density. Note that by the symmetric property of kernel functions, $K\left(\frac{\tau(x_i) - t_i}{h}\right) = K\left(\frac{t_i - \tau(x_i)}{h}\right)$: we use them interchangeably. By iterated expectation and the definition of conditional expectation:
\begin{align*} & \mathbb{E}\left[\frac{1}{h}K\left( \frac{t_i - \tau(x_i)}{h} \right)\frac{y_i}{Q_i}\right] = \mathbb{E} \left[\mathbb{E}  \left[\frac{1}{h}K\left( \frac{t_i - \tau(x_i)}{h} \right)\frac{y_i}{Q_i} \Big\vert x_i \right]\right] \\
&= \mathbb{E}\left[ \int \frac{y_i}{f_{T\mid X}(y_i\mid t') h} K\left(\frac{t' - \tau(x_i)}{h}\right) f_{Y,T\mid X}(y_i,t'\mid x_i ) dt'\text{ } dy \right]  \end{align*}
By a change of variables, let $ u = \frac{t' - \tau(x_i)}{h}$. Then $t' = hu + \tau(x_i)$ and $dt' = \frac{du}{h}$. Changing variables in the integral corresponds to computing a local expansion of the conditional outcome density $f_{Y \mid T, x}$ around $\tau(x_i)$.
\begin{align}
\mathbb{E}\left[ \int  y_i K(u) \frac{f_{Y,T\mid X}(y_i,  \tau(i)+hu \mid  x_i) }{f_{T\mid X}(y_i\mid \tau(x_i)+hu )} du \text{ } dy \right] = \mathbb{E}\left[ \int y_i K(u)  f_{Y\mid T, X}( y_i \mid hu + \tau(x_i), x_i) du \text{ } dy \right]
\label{lemma1}	\end{align}
We use the definition of Bayes' rule and conditional densities, $f_{Y,T \mid X} = f_{Y\mid T, X} f_{T\mid X}$, with the definition of Q as $f_{T\mid X}(t_i, x_i)$, to transform the conditional density from the conditional density to the target density, $f_{Y\mid T,X}$.

Consider a 2nd order Taylor expansion of $f_{Y\mid T, X}$ around $T = \tau(x_i)$:
\begin{align*}
& f_{Y\mid T, X}(
y_i\mid  \tau(x_i) + hu,x_i
) \\
& \approx f_{Y\mid T, X}(y_i\mid \tau(x_i),x_i) + hu\left(\frac{\partial}{\partial T} f_{Y\mid T, X}(y_i\mid \tau(x_i)) \right) + \frac{(hu)^2}{2} \frac{\partial^2}{\partial T^2} f_{Y\mid T, X}(y_i \mid \tau(x_i),x_i)  + o(h^2)
\end{align*}
Then we can compute the conditional expectation by integrating the approximation to the density term by term, where $\kappa_j(K) = \int u^j K(u) du$ denotes the jth kernel moment. The second order term describes the bias.
\begin{align*}
&  \mathbb{E}\left[\frac{1}{h}K\left( \frac{\tau(x_i) - t_i}{h} \right)\frac{y_i}{Q_i}\right] = \mathbb{E}\left[\int y_i K(u)  f_{Y\mid T, x} (y_i, \tau(x_i) + hu) du \text{ } dy \right] \text{ by eq.}\ref{lemma1} \\ & =   \mathbb{E}\left[ \int y_i K(u) f_{Y\mid T, x}(y_i, \tau(x_i)) du \text{ } dy\right] +   \mathbb{E}\left[\int y_i K(u) hu  \frac{\partial}{\partial T} f_{Y\mid T, x}(y_i, \tau(x_i)) du\text{ } dy\right]   \\
& +   \mathbb{E}\left[\int y_i K(u)(hu)^2 \frac{\partial^2}{\partial T^2}f_{Y\mid T, x}(y_i, \tau(x_i)) du\text{ } dy\right] +   \mathbb{E}\left[\int \frac{y_i}{n} o(h^2 )K(u)du \text{ } dy\right]\end{align*}
For a symmetric kernel, the odd-order moments integrate to 0.
\begin{align*}
& = \mathbb{E}\left[\int {y_i} f_{Y\mid T, x}(y_i \mid \tau(x_i)) dy \right] + \mathbb{E}\left[ \int \frac{1}{2}  {y_i} \frac{\partial^2}{\partial T^2} f_{Y\mid T, x}(y_i, \tau(x_i)) h^2 \kappa_2(K) dy \right] \\ &+ o(h^2)\int  y_i dy \text{          } \text{		 since } \int K(u) du = 1 \\
& = \mathbb{E}[Y(\tau(x_i))] +\kappa_2(K)  \mathbb{E}\left[ \int  \frac{y_i h^2 }{2} \frac{\partial^2}{\partial T^2} f_{Y\mid T, x}(y_i, \tau(x_i)) dy \right] + o(h^2)\int  y_i dy
\end{align*}
%

For the bias to vanish asymptotically, we require that $h^2 \to 0$, assuming that outcomes $y_i$ are bounded, and that the second derivative of the conditional density of $y$ given $T,X$ is bounded.

We also consider the multivariate case. We assume the kernel function for the vector $\textbf{u}$ is a product kernel, in the sense that it is the product of univariate kernels: $K(\textbf{u}) = \prod K(u_i)$, each with bandwidth $h_i$. The multidimensional change of variables takes the form $t = \tau(x_i) + H\textbf{u}$.  Then by the multivariate Taylor expansion of the conditional density $f_{Y \mid T,X}$, the bias is:

\[ \Bias(\tau) = \frac{\kappa_2(K)}{2} \sum_{j=1}^d \frac{\partial^2}{\partial T_j^2} f_{Y\mid T,X}(y_i\mid \tau(x_i),x_i)h_j^2 + o(\sum_{j=1}^d h_j^2) \]

%

\textbf{Calculations for Variance}:
A similar analysis follows for considering the variance; since we assume data $(X_i, Y_i)$ are i.i.d. it suffices to consider the varaince of one term of the estimator.
\begin{align*}
Var
\left[\sum_i^n \frac{1}{nh}\frac{y_i}{Q} K\left(\frac{t_i - \tau(x_i)}{h}\right)\right] =\frac{1}{nh^2} Var\left[ \frac{y_i}{Q} K\left(\frac{t_i - \tau(x_i)}{h}\right)\right] \\
= \frac{1}{n}\left( \mathbb{E}\left[ \left( \frac{y_i}{hQ} K\left(\frac{t_i- \tau(x_i)}{h}\right) \right)^2\right] -   \mathbb{E}\left[ \frac{y_i}{hQ}K\left( \frac{t_i - \tau(x_i)}{h}\right)\right]^2 \right)
\end{align*}

We will rewrite the squared expectation $ \left(\mathbb{E}\left[\frac{y_i}{h} K\left( \frac{t_i - \tau(x_i)}{h}\right) \right]\right)^2$ using the bias analysis as approximately  $ \left(\mathbb{E}[Y(\tau(X))] {h^2} \mathbb{E}\left[ \int  \frac{y_i}{2} \frac{\partial}{\partial T^2} f_{Y\mid T, x}(y_i\mid \tau(x_i),x_i)  \kappa_2(K) dy \right] + {o(h^2)} \int  y_i dy\right)^2$ for bounded outcomes $y_i$.

We compute $ \mathbb{E}\left[\left(\frac{y_i}{h} K\left( \frac{T - \tau(x_i)}{h}\right) \right)^2 \right]$ by analyzing a Taylor expansion of the conditional density after a change of variables:
\begin{align*}
& \mathbb{E}\left[\left( \frac{y_i}{hQ_i}  K\left(\frac{t_i- \tau(x_i)}{h}\right) \right)^2\right] = \mathbb{E}\left[ \int \int_{-\infty}^{\infty}  \frac{y_i^2}{h}K\left(\frac{t' - \tau(x_i)}{h}\right)^2 \frac{f_{Y,T}(y_i,t')}{f_{T \mid X}(\tau(x_i) + hu \mid x_i)^2} dt' dy \right]\\
& =  \mathbb{E}\left[\int \int_{-\infty}^{\infty}\frac{1}{h}\frac{y_i^2K(u)^2f_{Y\mid T, x_i}(y, \tau(x_i) +hu)}{f_{T \mid X}(\tau(x_i) + hu \mid x_i)}   du dy\right] \text{ by Bayes' rule}
\end{align*}
For convenience, we denote the quotient $\frac{f_{Y\mid T, x_i}(y, \tau(x_i) +hu,x_i)}{f_{T \mid X}(\tau(x_i) + hu\mid x_i)} = g_y(\tau(x_i) + hu)$. We expand this function around the argument $T$. In the asymptotic perspective, we omit the exact expressions of the derivatives $g', g''$, but we will require that the treatment density is nonzero for $\tau(x_i)$ (a standard overlap condition required for counterfactual policy evaluation).
Then consider each term of the expansion in turn: \begin{align*}
\mathbb{E}\left[\int \int_{-\infty}^{\infty}{y_i^2} K(u)^2 \left(g(\tau(x_i)) + g'(\tau(x_i))(hu) + (hu)^2 \frac{g''(\tau(x_i))}{2}\right)  du dy\right]
\end{align*}

The first term is equivalent to \[\frac{R(K)}{h}\mathbb{E}\left[\int \frac{  y_i^2 f_{Y \mid T, X} (y_i\mid \tau(x_i),x_i)  dy}{
f_{T \mid X}(\tau(x_i) \mid x_i)} \right] = \frac{R(K)}{h} \mathbb{E}\left[\frac{\mathbb{E}[Y_i^2 \mid \tau(x_i), X]}{f_{T \mid X}(\tau(x_i) \mid x_i)}\right]   \]

The second term and third terms are equivalently would require integrating $ \int u^2 K(u)du $ by parts, which requires specification on the structure of the kernel. Since the integration of the terms evaluate to constants, under the assumption that $h \to 0$, the integral of the second and third terms of the expansion is $O(h) = o(\frac{1}{h})$.

So \[\frac{1}{h}\mathbb{E}\left[\left( \frac{y_i}{Q}  K\left(\frac{t_i - \tau(x_i)}{h^2}\right) \right)^2\right] \approx \frac{R(K)}{h} \mathbb{E}\left[\frac{\mathbb{E}[Y^2 \mid \tau(X), X]}{f_{T \mid X}(\tau(X) \mid X)}  \right] + o(\frac{1}{h}) \]

$R(k)$ is the `roughness' term, where $R(k) = \int K^2(u) du $.

Combining these results:
\begin{align*}
& \frac{1}{nh^2} Var
\left[\frac{y_i}{Q} K\left(\frac{T - \tau(x_i)}{h}\right)\right]\\
&\approx  \frac{R(K)}{nh}\mathbb{E}\left[\frac{\mathbb{E}[Y^2 \mid \tau(X), X]}{f_{T \mid X}(\tau(X) \mid X)}  \right] + o\left(\frac{1}{nh}\right) - \frac{1}{n} \left( \mathbb{E}[Y(\tau(X))] + Bias \right)^2\\
&  \approx  \frac{R(K)}{nh} \mathbb{E}\left[\frac{\mathbb{E}[Y^2 \mid \tau(X), X]}{f_{T \mid X}(\tau(X) \mid X)}  \right]  + \frac{O(h^4) }{n} + o\left(\frac{1}{nh}\right)
\end{align*}

We discuss the multivariate case: again, we use the analysis of bias for the squared expectation term. For $ \mathbb{E}\left[\left(\frac{y_i}{\prod_j^d h_j } K\left( \frac{T - \tau(x_i)}{\prod_j^d h_j}\right) \right)^2 \right]$, we repeat the same analysis and use the product kernel form to decompose $K(\vec{u})^2 = \prod_{j=1}^d K(u_j)^2$.
Then\begin{align*}
&	\mathbb{E}\left[\int \int_{-\infty}^{\infty}{y_i^2} \prod_{j=1}^d K(u_j)^2 \left(g(\tau(x_i)) + \sum_{j=1}^d \frac{\partial}{\partial u_i }g(\tau(x_i))h_i u_i +\sum_{j=1}^d (h_iu_i)^2 \frac{\partial^2}{\partial u_i^2 }{g(\tau(x_i))}\right)  d\textbf{u} dy\right]   \\
&	= \frac{R(K)^d}{\prod_j^d h_j} \mathbb{E}\left[\frac{\mathbb{E}[Y_i^2 \mid \tau(x_i), X]}{f_{T \mid X}(\tau(x_i) \mid x_i)}\right] + O(\prod_j^d h_j)
\end{align*}

\end{proof}

\subsection{Analysis of Mean Squared Error: Proof of Theorem \ref{thm-mse}}\label{pf-mse}
\begin{proof}
Proof of Theorem \ref{thm-mse}:
We analyze the MSE and characterize the optimal bandwidth which minimizes the MSE.
\begin{align*}
& \mathbb{E}\left[\left( \frac{1}{n h} \sum_{i=1}^n K\left( \frac{\tau(x_i) - t_i}{h} \right)\frac{y_i}{Q_i} - \mathbb{E}[Y(\tau(X))]\right)^2\right] = (\Bias_\tau)^2 + \Var_\tau\\
= & (\Bias_\tau)^2 + \frac{R(K)}{nh} \mathbb{E}\left[\frac{\mathbb{E}[Y_i^2 \mid \tau(X), X]}{f_{T \mid X}(\tau(x_i) + hu \mid x_i)}\right] + o\left(\frac{1}{nh}\right) - \frac{1}{n} \left( \mathbb{E}[Y(\tau(X))] + \Bias_\tau \right)^2 \\
= & \frac{R(K)}{nh} \mathbb{E}\left[\frac{\mathbb{E}[Y_i^2 \mid \tau(X), X]}{f_{T \mid X}(\tau(x_i) + hu \mid x_i)}\right] + o\left(\frac{1}{nh}\right) +  O\left( \frac{h^4}{n} \right) + \frac{o(h^4)}{n} + O\left(\frac{1}{n}\right)
\end{align*}
For notational convenience, we denote the following constant factors:

$C_1 = \mathbb{E}\left[ \int  \frac{y_i}{2} \frac{\partial}{\partial T^2} f_{Y\mid T, x}(y_i, \tau(x_i))  \kappa_2(K) dy \right]$, $C_2 = \int  y_i dy$, $C_3 ={R(K)} \mathbb{E}\left[\frac{1}{Q}  \mathbb{E}[Y_i^2 \mid \tau(x_i), X]\right] $

If we want to optimize the bias-variance tradeoff of the asymptotic mean squared error, we choose the optimal bandwidth $h$ such that neither term dominates the other.
\begin{align*}
= & (Bias)^2 + Variance \approx  {h^4}\left(C_1^2 - O\left(\frac{1}{n}\right)\right)+ \frac{1}{nh} C_3  + o(h^4) (2  C_1 C_2+  C_2^2)
\end{align*}
Optimizing the leading terms of the asymptotic MSE with respect to the bandwidth $h$:
\begin{align*}
&\frac{d}{dh} (MSE) = 4 C_1^2 h^3 - \frac{C_3}{nh^2} = 0 \\
h^* = &\left(\frac{C_3}{4 C_1^2 n}\right)^{\frac{1}{5}} = \left(\frac{{R(K)} \mathbb{E}\left[\frac{\mathbb{E}[Y^2 \mid \tau(X), X]}{f_{T \mid X}(\tau(X), X)}\right]}{4 (\mathbb{E}\left[ \int  \frac{y_i}{2} \frac{\partial}{\partial T^2} f_{Y\mid T, x}(y_i, \tau(x_i))  \kappa_2(K) dy \right])^2 n}\right)^{\frac{1}{5}} = O\left(n^{-\frac{1}{5}}\right)
\end{align*}

The order of the optimal bandwidth is $O(\frac{1}{n^5})$; however in general, it will rely on the true density which is unknown a priori.
\end{proof}

\subsection{Analysis of consistency (Proof of Theorem \ref{thm-consistency})}\label{pf-consistency}

\begin{proof}

The convergence analysis of Theorem \ref{thm-mse} provides rate on the convergence of the estimator: at the optimal bandwidth $h = O\left(n^{-\frac{1}{5}} \right)$, and therefore the asymptotic terms $O\left(\frac{h^4}{n}\right)$ and $O\left(\frac{1}{nh} \right) = O(n^{-\frac{4}{5}})$ control the rate of convergence. Since the MSE decays at the rate $ O(n^{-\frac{4}{5}})$, the estimator converges in probability at the rate ${n}^{-\frac{2}{5}}$. This somewhat slower rate of convergence is standard in kernel density estimation.

For weak and strong uniform consistency results, we cite the results of \cite{s78} which hold if the density is uniformly continuous with bounded variation, and under additional moment conditions. In addition to absolute integrability of the kernel function, we require additionally that $\int \| x log (x) \|x \|^{\frac{1}{2}} \| dK(x) < \infty $. These conditions are satisfied by most kernel functions, including the Gaussian kernel.

Then, the empirical estimator converges almost surely. Theorem B of \cite{s78} provides exact rates of the convergence as well: $((nh)^{-1}log(\frac{1}{h}))^{\frac{1}{2}} \overset{a.s.}{\to} 0 $. With an optimal bandwidth of $h = O(n)^{-\frac{1}{5}}$, this corresponds to a $n^{\frac{2}{5}} \sqrt{ log(n)}$ rate for strong uniform consistency.
\end{proof}

\subsection{Analysis of consistency for normalized estimator}\label{pf-norm-consistency}
First we show that the normalization term converges to 1 in probability: $\hat{w}_n = \mathbb{E}\left[\sum\frac{1}{nhQ_i} K\left(\frac{\tau(x_i)-t}{h}\right) \right] \to 1$.

\[\mathbb{E}\left[\sum\frac{1}{nhQ_i} K\left(\frac{\tau(x_i)-t}{h}\right) \right] = \int \frac{1}{nhQ_i} K\left(\frac{\tau(x_i)-t}{h}\right) dt dy  \]
\[= \int \frac{f_{Y \mid T,X}(y\mid t, x)+o(h^2)}{f_{T\mid X}(t,x)}  dt dy = \int f_{Y,T\mid X}(y,t\mid x) dt dy +o(h^2) = 1  +o(h^2) \to 1\]

Then, by Slutsky's theorem, since each term in the quotient converges in probability and $\hat{w}_n \to_p 1$ which is a constant, $\hat{v}_{\tau}^{\text{norm}} \to V_\tau$.

\subsection{Consistency of Policy Optimization}\label{pf-opt}

Our generalization bound follows from direct application of the upper bound of Theorem 8 of \cite{bm2002}, the previous analysis characterizing the expectation of the estimator as  $ V_\tau= \hat{v}_\tau + \Bias(\tau)$, and the fact that $V_{\tau^*} \leq V_{\hat{\tau}} $:
\[ V_{\tau^*} - \mathrm{Bias}(\tau^*) = \Risk(\tau^*) \leq \Risk(\hat{\tau}_n) \leq V_{\hat{\tau}_n} - \mathrm{Bias}(\hat{\tau}_n)  \]

In our setting, since the cost function is assumed bounded with constant $\bar{M}_c$, there is an additional multiplicative factor on the concentration term $\sqrt{\frac{8 ln(2/\delta)}{n}}$.

The kernel function is bounded by definition and Lipschitz continuous depending on its specific functional form; we assume the Lipschitz constant of $K(u)$ is $L_K$, which is multiplied by $\frac{1}{h^2}$ in the empirical estimator. By Assumption~\ref{consasn5}, the inverse propensity weights are bounded $\frac{1}{Q_i}\in[1, \bar{M}_Q]$, $\forall x, T_i$.

The result follows from the Rademacher generalization bound \citep[Thm.~8]{bm2002}
concentrating $\E\hat v_\tau$ near $\E\hat v_\tau$, Thm.~\ref{thm-bias} relating $V_\tau$ to $\E\hat v_\tau$, and the Rademacher comparison lemma \citep[Thm.~4.12]{ledoux1991probability}.

\subsection{Gradient of estimator with Epanechnikov kernel for optimization}

For concreteness, we compute the gradient of our estimator for use in optimization algorithms for the Epanechnikov kernel. Note that the Epanechnikov kernel $K(u) = \frac{3}{4}(1 - u^2)\mathbb{I}\{\vert u \vert \leq 1\}$is nondifferentiable at the boundary where $u = 1$; we assume this happens with probability 0 on real data. The optimization problem is nonsmooth and nonconvex.

The gradient of the un-normalized linear policy estimator is:
\[ \nabla \hat{v}_{\tau} = \sum_{i=1}^n -x_i \frac{2(\beta^Tx_i-t_i)}{h}\frac{y_i}{Q_i} \mathbb{I}{\{\vert \frac{\beta^Tx_i-t_i}{h} \vert \leq 1 \}} \]

We can compute the gradient directly of the normalized estimator via the quotient rule: for simplicity, denote the sum of probability ratios as $\mathrm{norm}(Q)$, and denote the indicator function for each data point $\mathbb{I}\{\vert\frac{\beta^Tx_i - t_i}{h} \vert\leq 1 \} = \mathbb{I}(\beta^Tx_i)$.

\[ \nabla \hat{v}_\tau^{\mathrm{norm}} = \frac{-3}{2h} \frac{\left(\mathrm{norm}(Q) \left(\sum_i^n \frac{x_i y_i}{Q_i} \frac{\beta^Tx_i - t_i}{h} \mathbb{I}(\beta^Tx_i)\right)  -   {\hat{v}_\tau}\left( \sum_{i=1}^n \frac{x_i}{Q_i} \frac{\beta^Tx_i - t_i}{h} \mathbb{I}(\beta^Tx_i)\right)\right)}{\mathrm{norm}(Q)^2} \]

\subsection{Optimizing with the Triangular kernel} 

In the paper we reference the Epanechnikov kernel for concreteness and because it enjoys favorable properties such as statistical efficiency, as well as smoothness and concavity on its support. We remark that using the triangular kernel $K(u) = 1 - \vert u \vert, \vert u \vert \leq 1$ can allow us to rewrite the policy optimization problem (\ref{opt}) as a difference of convex programs, a special structural form of nonconvex optimization that can admit faster optimization by the Concave-Convex procedure (which alternatingly majorizes the concave portion). The global convergence of such a procedure remains an open question but it works well in practice \cite{cccp09}. The advantage of such a formulation is that vs. using numerical procedures for nonconvex optimization (BFGS) which have limited support for constraints, we can use modeling languages for convex solvers with richer support for constraints. 

Denote $\ell^h(a,u) = \max(a, u)$ as the hinge loss. Then the expression for the triangular kernel can be expressed as $K(u) = \ell^h(-1, u) - 2\ell^h(0, u) +l^h(1,u)$. Since the hinge loss is a convex function, we can write the objective as a difference of convex functions depending on the sign of $y_i$, which is data. The program remains convex if we optimize over a convex policy space.

\end{document}